\documentclass[letterpaper]{article} % DO NOT CHANGE THIS
\usepackage{aaai2026}  % DO NOT CHANGE THIS
\usepackage{times}  % DO NOT CHANGE THIS
\usepackage{helvet}  % DO NOT CHANGE THIS
\usepackage{courier}  % DO NOT CHANGE THIS
\usepackage[hyphens]{url}  % DO NOT CHANGE THIS
\usepackage{graphicx} % DO NOT CHANGE THIS
\usepackage{natbib}  % DO NOT CHANGE THIS AND DO NOT ADD ANY OPTIONS TO IT
\usepackage{caption} % DO NOT CHANGE THIS AND DO NOT ADD ANY OPTIONS TO IT
\usepackage{algorithm}
\usepackage{algorithmic}

\usepackage{amsmath, mathrsfs}
\usepackage{booktabs}
\usepackage{cancel}
\usepackage{xcolor}

\newcommand{\up}[2]{$\text{#1}_{\color{red}{\text{+#2}}}$}
\newcommand{\down}[2]{$\text{#1}_{\color{blue}{\text{-#2}}}$}
\newcommand{\titleabb}{SamKV}
\newcommand{\titleabbs}{SamKV }
\usepackage{amssymb}

\usepackage{colortbl, booktabs}
\usepackage{array}
\usepackage{pifont}

\usepackage{utfsym}
\newcommand{\go}{\usym{1F5F8}}
\newcommand{\cha}{\usym{2613}}

\usepackage{newfloat}
\usepackage{listings}
\DeclareCaptionStyle{ruled}{labelfont=normalfont,labelsep=colon,strut=off} % DO NOT CHANGE THIS
\floatstyle{ruled}
\newfloat{listing}{tb}{lst}{}
\floatname{listing}{Listing}
\title{Sparse Attention across Multiple-context KV Cache}
\author{
    Ziyi Cao\textsuperscript{\rm 1,}\thanks{Work done during the internship at Huawei.}\textsuperscript{\rm ,}\equalcontrib,
    Qingyi Si\textsuperscript{\rm 2,}\equalcontrib,
    Jingbin Zhang\textsuperscript{\rm 2,}\equalcontrib,
    Bingquan Liu\textsuperscript{\rm 1}
}
\affiliations{
    \textsuperscript{\rm 1}Harbin Institute of Technology\\
    \textsuperscript{\rm 2}Huawei Technologies Co., Ltd.
    zyc@stu.hit.edu.cn,
    siqingyi@huawei.com,
    zhangjingbin@pku.edu.cn,
    liubq@hit.edu.cn
}

\usepackage{bibentry}
\begin{document}

\maketitle

\begin{abstract}
Large language models face significant cost challenges in long-sequence inference. To address this, reusing historical Key-Value (KV) Cache for improved inference efficiency has become a mainstream approach. Recent advances further enhance throughput by sparse attention mechanisms to select the most relevant KV Cache, thereby reducing sequence length. However, such techniques are limited to single-context scenarios, where historical KV Cache is computed sequentially with causal-attention dependencies. In retrieval-augmented generation (RAG) scenarios, where retrieved documents as context are unknown beforehand, each document’s KV Cache is computed and stored independently (termed multiple-context KV Cache), lacking cross-attention between contexts. This renders existing methods ineffective. Although prior work partially recomputes multiple-context KV Cache to mitigate accuracy loss from missing cross-attention, it requires retaining all KV Cache throughout, failing to reduce memory overhead.
This paper presents \textbf{\titleabb}, the first exploration of attention sparsification for multiple-context KV Cache. Specifically, \titleabbs takes into account the complementary information of other contexts when sparsifying one context, and then locally recomputes the sparsified information. Experiments demonstrate that our method compresses sequence length to 15\% without accuracy degradation compared with full-recompuation baselines, significantly boosting throughput in multi-context RAG scenarios. 
\end{abstract}


\section{Introduction}
Large language models (LLMs) \citep{DBLP:conf/nips/VaswaniSPUJGKP17,DBLP:conf/naacl/DevlinCLT19} have demonstrated remarkable capabilities across a multitude of domains, including question answering, chatbots, education, and healthcare.
They process text-like token sequences submitted by users to provide responses.
However, as requests (user query) become more intricate, the challenge of serving LLMs effectively grows more pronounced.
This is particularly evident in complex tasks such as multi-context question answering and few-shot learning within retrieval-augmented generation (RAG) scenarios \citep{DBLP:conf/nips/LewisPPPKGKLYR020}.
These tasks often involve immutable token chunks (e.g., system messages, examples, and contexts) that remain unchanged and are repeatedly used across different requests.
The key challenge lies in how to effectively leverage these recurring text chunks to enhance the efficiency of LLM services.

\begin{figure}[t]
    \centering
    \includegraphics[width=0.8\linewidth]{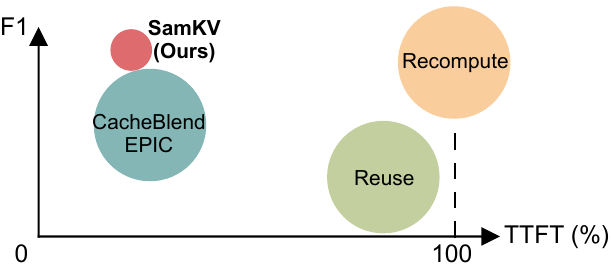}
    \caption{KV Cache methods across multiple contexts. The x-axis shows time to first token (TTFT) as a percentage of full recomputation. The y-axis represents F1 scores, with circle sizes indicating GPU memory usage during inference.}
    \label{fig:first}
\end{figure}

\begin{table}[t]
\small
    \centering
    \begin{tabular}{lccc}
    \toprule
       Multi-context methods & CacheBlend & EPIC & Ours \\
    \midrule
       Sequence ratio & 100\% & 100\% & 14.9\% \\ 
       Recomputation ratio & 15.0\% & 14.1\% & 14.3\% \\   
    \bottomrule
    \end{tabular}
    \caption{Sequence ratio indicates the proportion of KV Cache that requires loading in the GPU, and recomputation ratio indicates the ratio of tokens to be recomputed.}
    \label{tab:compare}
\end{table}

To address the challenges of LLM serving, context caching has emerged as a key strategy to enhance efficiency by reusing precomputed Key-Value (KV) Caches \citep{DBLP:conf/mlsys/PopeDCDBHXAD23} for repeated tokens. Context caching methods can be broadly categorized into two types: single-context and multi-context approaches. Single-context methods often concentrate on sparsifying the KV Cache, aiming to reduce the computational cost and memory footprint associated with LLM inference \citep{DBLP:conf/nips/LiHYVLYCLC24, DBLP:journals/corr/abs-2409-10516, DBLP:conf/nips/XiaoZ0XLZ0024}. However, these single-context approaches are not designed to address the absence of cross-attention that occurs when multiple contexts are prefilled separately. In contrast, multi-context techniques typically focus on recomputing specific tokens' KV Caches \citep{DBLP:journals/corr/abs-2405-16444,DBLP:journals/corr/abs-2410-15332}. This approach can somewhat alleviate the lack of cross-attention between different text chunks. Yet, it comes at a cost. These methods generally require the entire KV Cache to be loaded into memory, even if only a fraction of the tokens' KV Caches are recomputed. This means that while they improve the accuracy of the generated responses by addressing cross-attention issues, they do not sparsify the KV Cache. As a result, the GPU memory usage remains high, posing a challenge for efficient LLM serving, especially when dealing with extensive contextual information.

To break through these limitations, we introduce \titleabb, a novel method that combines sparsification and selective recomputation to efficiently handle multiple contexts. Our approach consists of two main steps. First, we perform sparsification by filtering out irrelevant content from the current document based on the user query and other documents' information. This step ensures that only the most useful content is retained, significantly reducing the amount of data to be processed. Second, we selectively recompute the KV Caches for a subset of tokens within the sparsified content. This selective recomputation focuses on updating the KV Caches for tokens that are crucial for maintaining cross-attention and generating high-quality responses. By integrating sparsification and selective recomputation, \titleabbs achieves both acceleration and reduced memory usage, making it more efficient to handle multiple contexts without the need to load the entire KV Cache into memory.

% Specifically, during sparsification, \titleabbs only uses the KV Cache from initial and local positions within a sequence. It generates a universal query vector based on the user's query, which is then used to incorporate query information from other contexts into the sparsification process of a given context. This ensures that the sparsification is not only relevant to the user's query but also context-aware. In the recomputation phase following sparsification, \titleabbs emphasizes recomputing the KV Caches for initial and local tokens. In addition to this, it selects a very small number of important tokens for recomputation. The selection of these tokens is based on their inherent importance within the context, ensuring that the most significant information is updated.

Specifically, during the sparsification process, \titleabbs exclusively utilizes the KV Cache from initial and local positions in the sequence to generate query vectors. To enhance cross-document information extraction, we further enrich each context’s query vector by incorporating query-relevant signals from other contexts. For instance, when answering a crime-related question, \titleabbs fetches the KV Cache of semantically linked documents (e.g., legal statutes and jurisprudential analyses). By augmenting the query vector for jurisprudential analysis with features from legal statutes, the system more effectively captures their inter-document correlations, enabling personalized query embeddings per context.
Moreover, during token recomputation, \titleabbs selectively processes only the sparsified KV Cache, prioritizing tokens from initial/local positions and those with high attention scores. This dual strategy reduces computational overhead while preserving model accuracy.

In this paper, we introduce \titleabb, a novel approach that achieves significant compression of multi-context KV Caches without compromising accuracy. Specifically, \titleabbs reduces the KV Cache size to only 15\% of the original while maintaining the same level of precision. Furthermore, we find that incorporating information from other contexts into the user query during the sparsification process yields more critical sparse information. Additionally, we observe that integrating the newly recomputed KV values for the sparse tokens back into the original KV Cache can sometimes enhance the overall performance.

Our contributions are primarily as follows:
\begin{itemize}
    \item We present the first approach to achieve KV Cache sparsification in multi-context scenarios, which significantly reduces GPU memory usage.
    % \item We propose an innovative sparsification method for multi-context attention that effectively compresses KV Caches without reducing accuracy.
    \item We propose an innovative multi-context sparsification approach and a novel token recomputation method, both of which achieve promising results.
    % \item We employ two fusion strategies: incorporating inter-context correlations during sparsification and blending new and old KV values during recomputation, both of which yield promising results.
    \item Our method, validated on the LongBench's Question-Answering (QA) datasets \citep{DBLP:conf/acl/BaiLZL0HDLZHDTL24}, demonstrates the ability to reduce computational load and accelerate processing while maintaining accuracy.
\end{itemize}

\section{Related work}

\subsection{Single-context sparsification}
Optimizing LLMs for long-context processing has focused on single-context sparsification techniques. For example, InfLLM \citep{DBLP:conf/nips/XiaoZ0XLZ0024} uses sliding window attention with an efficient context memory, while SnapKV \citep{DBLP:conf/nips/LiHYVLYCLC24} compresses the KV Cache by identifying critical attention features. However, these methods are designed for single contexts and do not address cross-attention issues in multi-context settings. In contrast, \titleabbs leverages the correlations between multiple contexts, incorporating cross-attention mechanisms to capture dependencies between contexts. When processing a single context, \titleabbs degrades gracefully to traditional single-context methods, making it a generalization of existing approaches.

\subsection{Multi-context recomputation}
Multi-context methods represent an extension of single-context approaches, further enhancing the efficiency and scalability of LLM serving. Recent works such as CacheBlend \citep{DBLP:journals/corr/abs-2405-16444} and EPIC \citep{DBLP:journals/corr/abs-2410-15332} have made significant strides in this area. CacheBlend selectively recomputes tokens based on the absence of cross-attention in the first layer, with the scope of updates decreasing progressively across layers. EPIC focuses on updating only the initial and local positions within the context to minimize computational load. In contrast, our proposed method, \titleabb, leverages the inherent importance of certain tokens within the context, identified through the concentration of attention weights (as observed in the RaaS \citep{hu2025raas}). By selectively recomputing these critical tokens, in addition to initial and local positions, \titleabbs reduces both the amount of KV Cache needed and the overall computational requirements, offering a more efficient and targeted approach to LLM serving.

\subsection{Attention sink}

In the quest to optimize LLMs, recent studies have uncovered distinct attention patterns. StreamingLLM \citep{DBLP:conf/iclr/XiaoTCHL24}  first identifies attention concentration at initial positions. Longformer \citep{DBLP:journals/corr/abs-2004-05150} emphasizes local contexts, highlighting the importance of nearby tokens. RaaS \citep{hu2025raas} further reveals attention clustering at intermediate positions during reasoning tasks. In contrast, \titleabbs extends these findings by considering both initial and local tokens and refining the threshold-based selection from RaaS. Using power-law distribution fitting and comparison, \titleabbs more effectively identifies critical tokens, offering a more rational and comprehensive approach to token selection for enhanced efficiency and performance.

\begin{table}[th]
\small
    \centering
    \begin{tabular}{cl}
    \toprule
       Notation & Description \\
    \midrule
       $N$ & total number of layers \\
       $N^*$ & stable attention layers (Appendix A.2) \\
       $D$ & total number of documents \\
       ${KV}_\text{doc-i}$ & document i's KV Cache (also for Q and K Caches)\\  % 同理适用于Q Cache
       $\text{doc-i}^\text{old}$ & $KV_{\text{doc-i}^\text{old}}$ represents the $KV_\text{doc-i}$ before recomputa- \\ & tion (also for new), defaulting to old \\
       $\text{doc-i}_\text{ini}$ & $\underline{\text{ini}}\text{tial}$ position of document i's absolute position \\ & (also for $\underline{\text{mid}}\text{dle}$, $\underline{\text{loc}}\text{al}$, or attention degree posi- \\ 
       & tions $\underline{\text{anc}}\text{hor}$, $\underline{\text{max}}$, and $\underline{\text{min}}$) (attention analysis \\ & is in Appendix A.1) \\
       $Q_\text{que}$ & generic query vector \\
       $\hat{Q}_\text{doc-i}$ & query vector for document i only \\
       $P_\text{doc-i}$ & select Top P in document i \\
       $\left<\cdot, \cdot\right>$ & inner product \\
       $KV^{(n)}$ & $n$-th layer of $KV$ (also for percentages) \\
       % $K_{\text{doc-i}_\text{anc}}$ & anchor point $K$ in document $i$  \\
       % $K_{\text{doc-i}_\text{max}}$ & max attention $K$ in document $i$ \\
       % $K_{\text{doc-i}_\text{min}}$ & min attention $K$ in document $i$ \\
       $s^{(n)}_{\text{doc-i}_\text{anc}}$ & $\left<Q^{(n)}_\text{doc-i}, K^{(n)}_{\text{doc-i}_\text{anc}}\right>$ (also for max and min) \\
    \bottomrule
    \end{tabular}
    \caption{Summary of terminology.}
    \label{tab:symbol}
\end{table}

\begin{figure}[ht]
    \centering
    \includegraphics[width=0.85\linewidth]{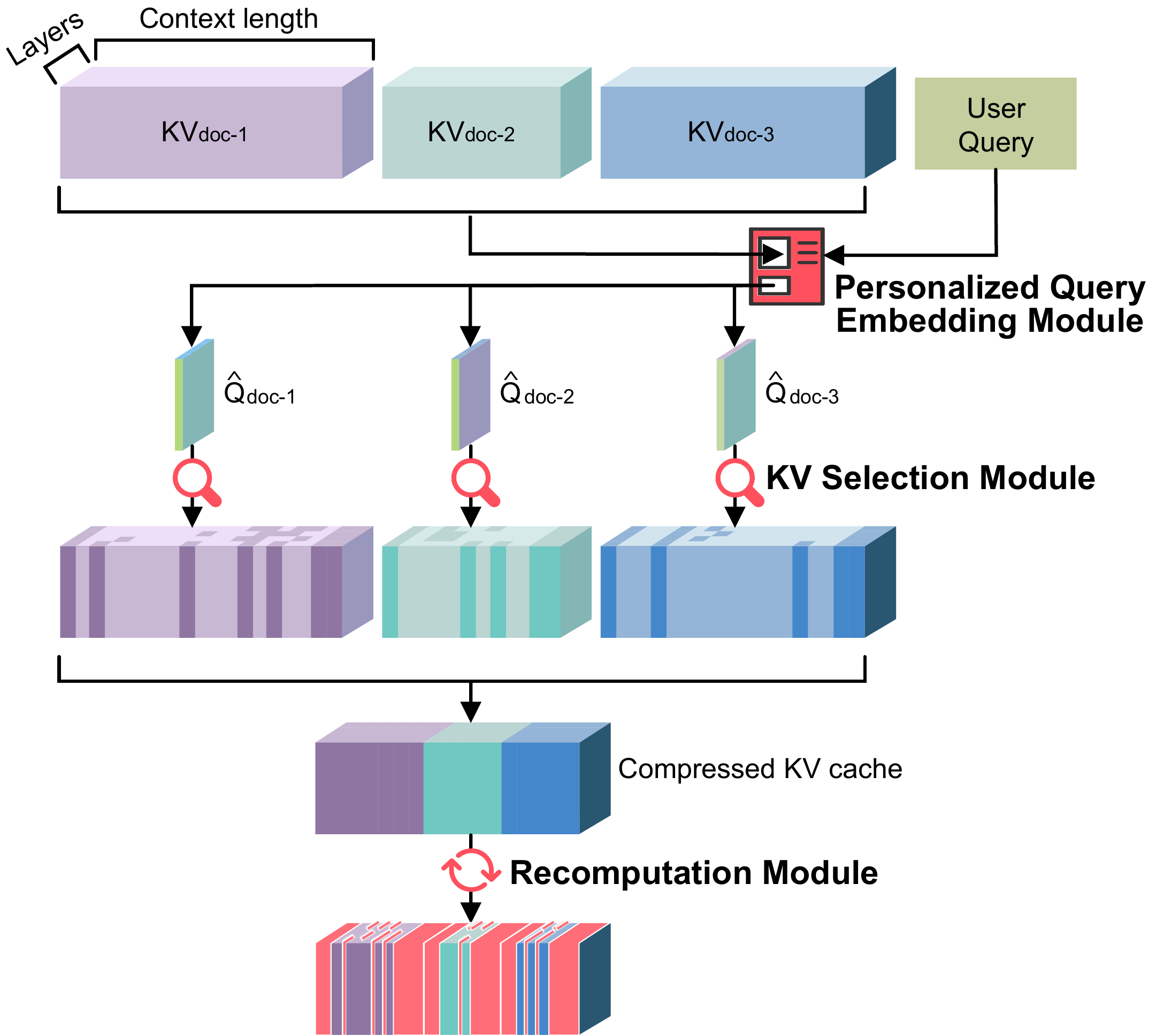}
    \caption{Overview of \titleabb: input user query text; generate exclusive personalized query vector for each context; get compressed KV Cache; recompute partial tokens.}
    \label{fig:overview}
\end{figure}

\begin{figure}[ht]
    \centering
    \includegraphics[width=0.9\linewidth]{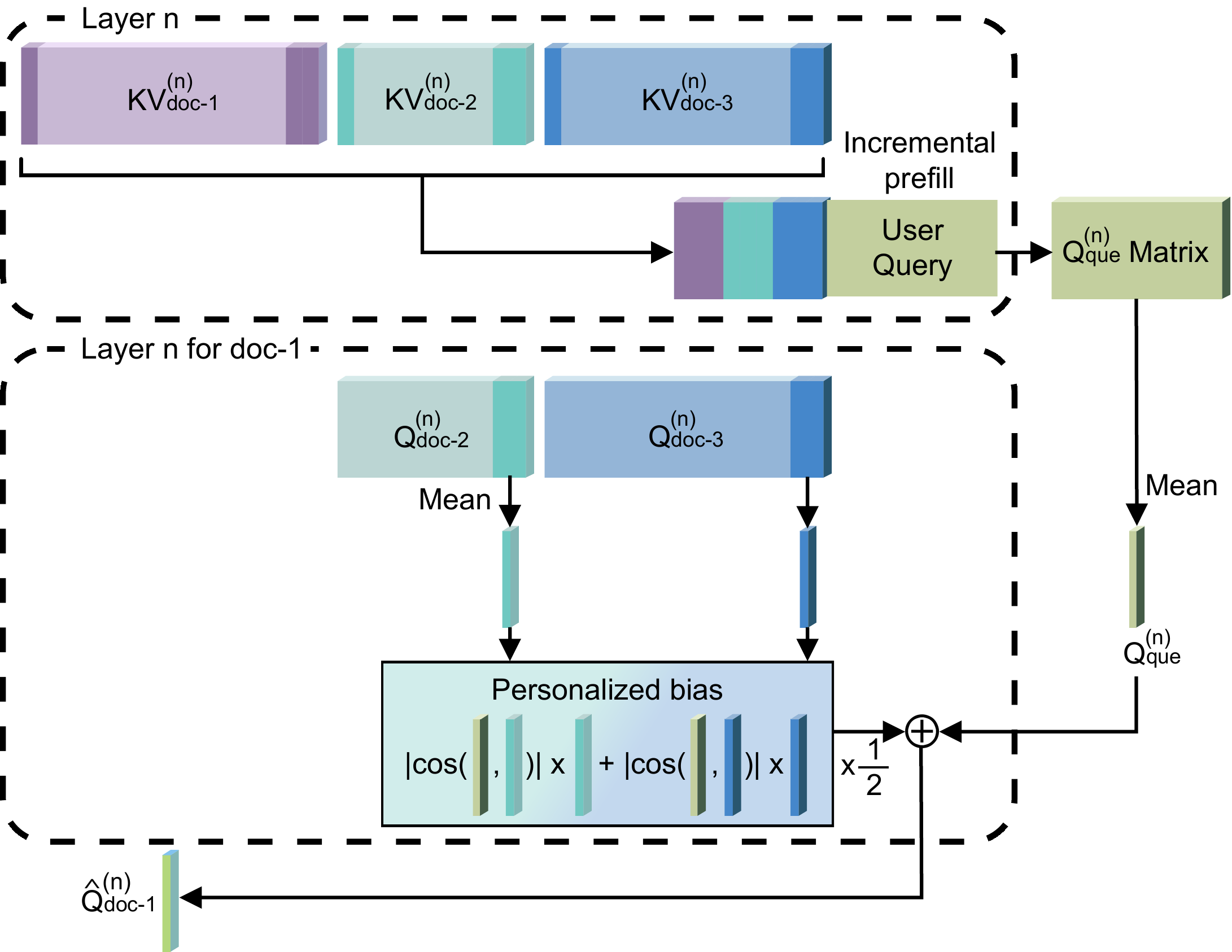}
    \caption{Detail of Personalized Query Embedding Module. The upper half depicts incremental prefill generating the generic query vector $Q_\text{que}$, while the lower half visualizes the integration of personalized bias (derived from $Q_{\text{doc-2}_\text{loc}}$ and $Q_{\text{doc-3}_\text{loc}}$) into $Q_\text{que}$ to form the document-specific $\hat{Q}_\text{doc-1}$.}
    % 实际代码中，虽然是用hidden states乘Q矩阵得到的，但是在没有bias的情况下，相当于直接提取Q states，为了简化表达，避免显得太复杂不易理解，这里直接将这个抽出来，本质差别不大
    \label{fig:query_vector}
\end{figure}

\section{Our proposed \titleabb}

When handling multiple contextual scenarios, two critical challenges emerge: efficiently sparsifying the KV Caches across contexts, and addressing the cross-attention deficiency caused by independent prefilling of each context. To resolve these issues, we propose \titleabb, a novel method designed to jointly optimize sparse KV Cache management and cross-context attention recovery.
\titleabbs comprises three core components: (i) Personalized Query Embedding Module, (ii) KV Selection Module, and (iii) Recomputation Module.
To elucidate the interplay between these modules, Figure~\ref{fig:overview} presents a running example with three distinct contexts.
The remainder of this section details each component's design and its role in \titleabb's pipeline.

\subsection{Generation of personalized query vectors}
\label{sec:query_vector}

% 显示为什么要生成，简单两句：因为现在的占用过高，要稀疏化
Existing approaches for multi-context KV Cache necessitate full cache loading \citep{DBLP:journals/corr/abs-2405-16444, DBLP:journals/corr/abs-2410-15332}, which incurs substantial GPU memory overhead. The development of sparse KV Cache techniques provides an effective alternative, where constructing appropriate query vectors represents the crucial initial phase in implementing such sparse solutions.
Unlike traditional approaches that employ incremental prefill for user queries, our method introduces document-specific query vector customization, enabling more efficient context-aware processing.

%… 【后置到重算部分，而不是放在这里】【已完成】这里还少一部分内容 ：这里可以先说一句‘multi-context场景下，query的q、各个文档的kv均来自于不同的prefill，缺少cross-attention，因此，直接用query的q去每个文档的k中检索是很难找到最相关的k片段的。为了解决这个问题，Personalized Query Embedding Module为每个文档专门定制化q vector，以实现在每个文档中分别找到最相关的片段。

% 计划：在开头出讲故事：为什么要加个性化偏置：因为由query得到的多上下文间是存在相同部分的，而这种共性正是很重要的，而添加其他文档的信息是有利于得到这部分的，所以我们要在原q上添加其他文档的q（将多文档不同轮次prefill问题改为这个，可能更好捋顺）
% 在具体添加时，直接讨论添加什么（local）

% In the multi-context scenario, the user query and the KV Cache of each document originate from distinct prefill operations, lacking cross-attention. Consequently, directly searching for the most relevant key fragments within each context using the generic user query is challenging. To address this issue, the Personalized Query Embedding Module customizes a dedicated vector for each document, enabling the identification of the most relevant segments within each document separately.
% Generation of personalized query vectors across multiple contexts is a two-step task: (i) generating the generic query vector, and (ii) attaching a personalized bias to it.
% 然后是具体的讲故事，通用的就不讲了（通用的理由就是为什么要制作查询向量的理由），而是将为什么要添加个性化偏置，因为有文档间共识的存在，只有添加上其他上下文信息才能更好地识别这部分信息。

Specifically, multiple contexts and their corresponding KV Caches can be firstly retrieved using the user query in the RAG scenario.
Since these contexts are all relevant to the query, they often overlap with each other to some extend.
We refer to this overlap as \textbf{inter-document consensus}.
The consensus, due to its repeated appearance across multiple documents, indicates its particular significance.
However, merely using the user query for sparsifying a context is not sufficient to effectively identify the consensus.
Therefore, when sparsifying a context, we also need to consider other contexts as query information to recognize the consensus.
This results in varying query vectors for each context.
To this end, we propose the Personalized Query Embedding Module, which first generates a generic query vector based on the user query information and then appends consensus information specific to the current context.
The details are as follows:

\textbf{Generate generic query vector.}
Inspired by the observation that KV Caches at both initial and local positions play a critical role in attention computation \citep{DBLP:conf/iclr/XiaoTCHL24, DBLP:journals/corr/abs-2004-05150}, we propose a novel cache compress strategy.
Specifically, we extract and concatenate the KV pairs from these two pivotal positions across all contexts to form a composite Cache unit.
This compressed Cache is then employed to incrementally prefill the Q Matrix of user queries through attention computation.
Here, the Q Matrix encapsulates the current contextual query representation, which is subsequently transformed into a spares query vector $Q_\text{que}$ via mean pooling. As illustrated in Figure~\ref{fig:query_vector}, the $Q_\text{que}^{(n)}$ denotes the $Q_\text{que}$ of the $n$-th layer.
% This design achieves dual objectives: (i) preserving essential positional information through strategic KV Cache selection, and (ii) enabling efficient query processing through dimensionality reduction.
% incremental prefill 得到user query的Q矩阵
% To reduce GPU memory, \titleabbs fetches only the important initial and local locations of the KV Cache involved in query vector generation.
% 添加上具体的描述，Q是根据前面的initial和local去生成
% As shown in the LLM running in Figure~\ref{fig:generic_query_vector}, the generic query matrix is extracted from the Q Matrix of the user query at the current layer, and the $Q$ is obtained by taking its mean in the sequence length dimension.

% 符号用下标做来源区分
% Q_{doc 1} Cache

% 图里不要15%single
% 表格里直接加infllm
% 图里改为multi-context infllm

% 加doc
\textbf{Add personalized bias.}
Although $Q_\text{que}$ is generated from context, its primary role is to express the user query, which limits its ability to capture targeted consensus.
To address this limitation, we propose augmentaing $Q_\text{que}$ with a contextual Q Cache.
As before, we focus on Q Caches at the initial and local positions.
The initial Q Cache is designed to interact with initial-position K, while the local Q Cache multiplies with all K to complete the information extraction.
Consequently, the local Q Cache exhibits the strongest retrieval capability, and we denote document i's local Q Cache as $Q_{\text{doc-i}_\text{loc}}$.
The next step is determining the proportion and method of integrating $Q_{\text{doc-i}_\text{loc}}$ into $Q_\text{que}$.
To avoid overshadowing the user's original query, the bias introduced by $Q_{\text{doc-i}_\text{loc}}$ must be weighted lightly.
Empirical analysis reveals that the cosine similarity between $Q_\text{que}$ and $Q_{\text{doc-i}_\text{loc}}$ typically ranges from -0.3 to 0.3.
Critically, only when $Q_{\text{doc-i}_\text{loc}}$ is incorporated into $Q_\text{que}$ with positive weighting does the inner product of $Q_\text{que}$ and any $K$ vector effectively retain the multiplicative interaction between $Q_{\text{doc-i}_\text{loc}}$ and $K$.
We therefore adopt the absolute cosine value $|\cos|$ as weight constraint.
To further safeguard against query information dilution from excessive context, we normalize the integration by the number of contextual documents.
The equation composition is expressed as:
\begin{align}
\small
\begin{aligned}
    \hat{Q}_\text{doc-i}=Q_\text{que}&+\frac{1}{D-1}\times\\
    &\sum_{\text{j} (\text{j}\neq\text{i})}^D |\cos(Q_\text{que}, Q_{\text{doc-j}_\text{loc}})|\times Q_{\text{doc-j}_\text{loc}}
    \end{aligned}
    \label{equ:q_doc_i}
\end{align}
This equation fundamentally ensures that when sparsifying a document, its query vector assimilates relevant query information from other contexts to facilitate consensus recognition.
As illustrated in Figure~\ref{fig:query_vector}, $\hat{Q}_\text{doc-1}$ for context 1 is synthesized through Equation~\ref{equ:q_doc_i} by combining $Q_\text{que}$ with both $Q_{\text{doc-2}_\text{loc}}$ and $Q_{\text{doc-3}_\text{loc}}$ according to the specified fusion scheme.
The vectors $\hat{Q}_\text{doc-2}$ and $\hat{Q}_\text{doc-3}$ are derived analogously, following the same augmentation process.

\subsection{Selection of important KV Caches}
\label{sec:select}

\begin{figure}[t]
    \centering
    \includegraphics[width=0.6\linewidth]{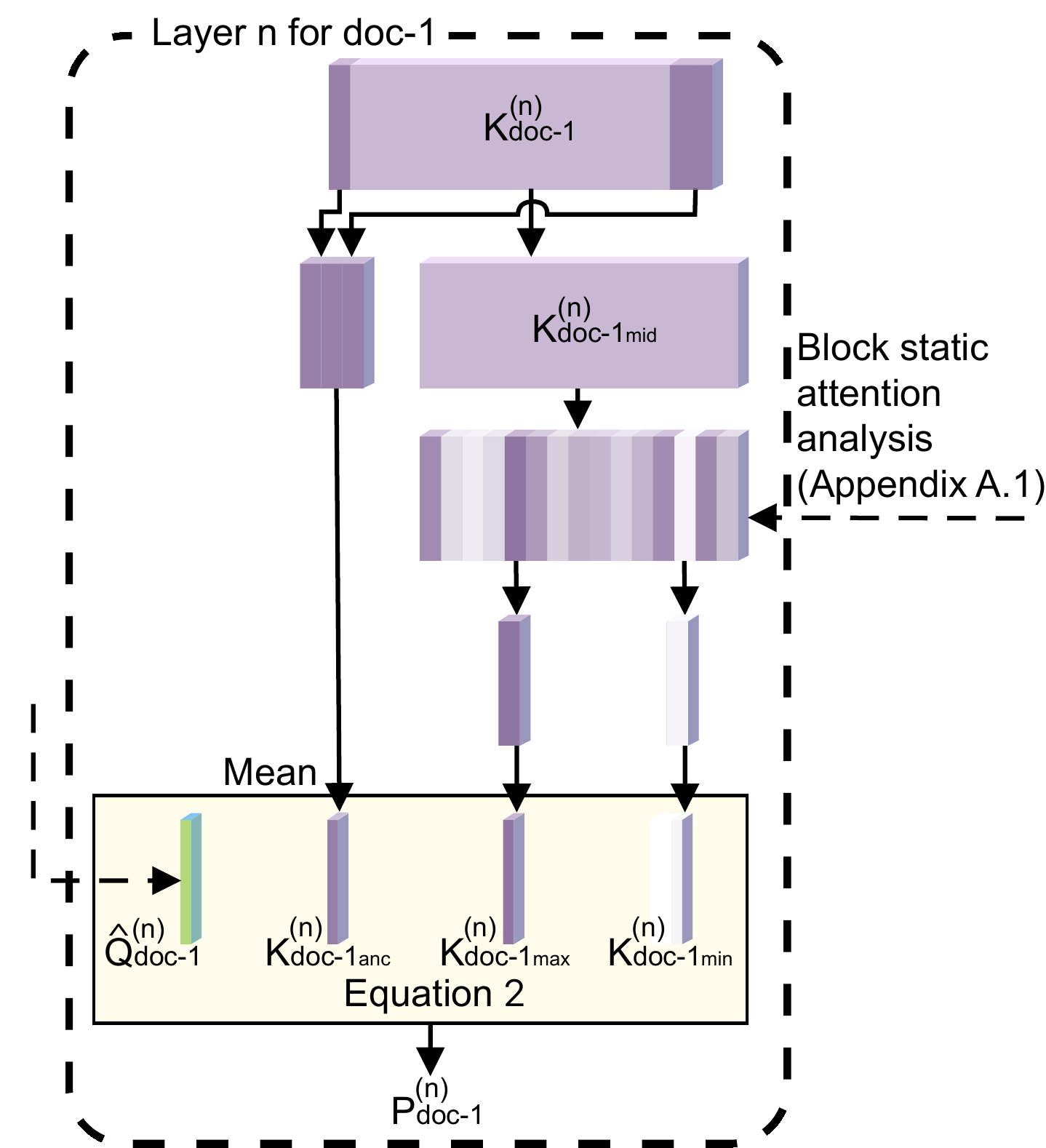}
    \caption{Calculate the Top P to be taken for context 1 in layer $n$.}
    \label{fig:get_p}
\end{figure}

% 不用冒号，用双重下标

Current multi-context methods require loading the entire KV Caches during inference, leading to excessive GPU memory consumption. To sparsify the multi-context KV Caches, beyond the query vector generation in Section~\ref{sec:query_vector}, another critical challenge is determining how many KV Caches to select. In our proposed \titleabb, we address this by introducing a dynamic Top-P sampling strategy based on anchor points: the proportion P scales adaptively with the number of K Caches more important than the anchor. 

Specifically, since KV Caches at initial and local positions are widely recognized as critical for inference \citep{DBLP:conf/iclr/XiaoTCHL24, DBLP:journals/corr/abs-2004-05150}, we retain them in full resolution without sparsification.
The primary target of sparsification is instead the middle segments of the KV Caches.
Given the importance of initial and local KV Caches, we leverage their corresponding K Caches as anchor points $K_{\text{doc-i}_\text{anc}}$ for attention pattern analysis.
For the middle blocks, we identify the blocks with maximum and minimum attention scores, denoted as $K_{\text{doc-i}_\text{max}}$ and $K_{\text{doc-i}_\text{min}}$ respectively (see Appendix A.1). It is important to note that some tokens (e.g., punctuation marks) may yield high attention scores while being semantically insignificant. To address this issue, we implement block-level KV management and retrieval, where each block is represented by the mean vector of its constituent token caches.

% \czy{For the middle blocks, we identify regions with the highest $K_{\text{doc-i}_\text{max}}$ and lowest $K_{\text{doc-i}_\text{min}}$ attention scores  but mitigate potential noise from outlier tokens, such as those with high attention weights but low semantic relevance, by aggregating each block via mean pooling to derive robust feature representation.}

Furthermore, we compute $K_{\text{doc-i}_\text{anc}}$, $K_{\text{doc-i}_\text{max}}$, and $K_{\text{doc-i}_\text{min}}$ inner products with $\hat{Q}_{\text{doc-i}}$ to obtain $s_{\text{doc-i}_\text{anc}}$, $s_{\text{doc-i}_\text{max}}$, and $s_{\text{doc-i}_\text{min}}$, respectively. Under the assumption that $s_{\text{doc-i}_\text{anc}}$ and $s_{\text{doc-i}_\text{min}}$ remain stable, a larger $s_{\text{doc-i}_\text{max}}$ (i.e., a higher upper bound of attention scores) implies more K Caches surpassing the anchor’s importance, thus $P_\text{doc-i}$ should increase proportionally with $s_{\text{doc-i}_\text{max}}$. By similar reasoning, $P_\text{doc-i}$ is inversely proportional to both $s_{\text{doc-i}_\text{min}}$ and $s_{\text{doc-i}_\text{anc}}$.
This principled analysis leads to our formulation in Equation~\ref{equ:pi}, i.e.,
\begin{align}
\small
    P^{(n)}_\text{doc-i}=\left\{\begin{aligned}
        &\frac{s^{(n)}_{\text{doc-i}_\text{max}} - s^{(n)}_{\text{doc-i}_\text{anc}}}{s^{(n)}_{\text{doc-i}_\text{max}}-s^{(n)}_{\text{doc-i}_\text{min}}}, \\
        &\qquad \text{ if }s^{(n)}_{\text{doc-i}_\text{anc}} \in \left(s^{(n)}_{\text{doc-i}_\text{min}}, s^{(n)}_{\text{doc-i}_\text{max}}\right], \\
        & 0,\quad\; \text{otherwise.}
    \end{aligned}\right.
    \label{equ:pi}
\end{align}

% Current multi-context methods necessitate carrying the entire KV Cache, incurring excessive GPU memory usage. Thus, there is an urgent need to sparsify by selecting the most important KV pairs from the multi-context KV Cache.
% Apart from the query vector $Q_i$, the other heart is in which part of the KVs to select how many KVs.
% Given the importance of KVs in the initial and local locations, only the middle part of the cache needs to be filtered, as seen in Figure~\ref{fig:get_p}\footnote{See Appendix A.1 for the importance analysis of the middle part.}.
% In calculating the selection percentage $P^{(n)}_i$, \titleabbs employs in a similar way to normalization, i.e., choosing anchor, maximum and minimum values.
% With $Q^{(n)}_i$ as the query feature, the anchor value corresponding to $K^{(n)}_{i:anc}$ denotes the relatively important K, while the maximum value corresponding to $K^{(n)}_{i:max}$ denotes the most important K in middle part.
% $K^{(n)}_{i:min}$ is defined similarly.
% Thus,

% 改表达
As formulated in Equation~\ref{equ:pi}, only those K Caches that surpass the anchor point in importance are selected as significant features.
While Equation~\ref{equ:pi} determines the selection ratio $P^{(n)}_\text{doc-i}$ specifically for layer $n$, the final ratio $P_\text{doc-i}$ is obtained by consolidating layer-wise ratios through Equation~\ref{equ:p}, i.e., 
\begin{align}
\small
    P_\text{doc-i}=\frac{1}{N^*}\sum_n^{N^*} P_{\text{doc-i}}^{(n)}.
    \label{equ:p}
\end{align}
In Equation~\ref{equ:p}, we selectively employ attention layers that exhibit stable token-wise attention scores, denoted as $N^*$. The derivation process and criteria for identifying $N^*$ are elaborated in Appendix A.2.

After selecting the Top $P_\text{doc-i}$ KV Cache blocks for document i, we normalize the inner products of all retrieved blocks across contexts before concatenating and comparing them. From this combined set, we retain only the most critical blocks, with the number of selected blocks equal to the total block count divided by the number of contexts. This ensures cross-contextual filtering of the most important blocks.
% 这里暂时注释掉了，篇幅不够用了
%A key challenge arises because each $Q_\text{doc-i}$ generates distinct inner product distributions, making direct comparison across documents impossible without normalization. Consequently, our method first normalizes these values before performing inter-document selection, rather than attempting to reduce $P_\text{doc-i}$ directly. The final output concatenates all filtered KV Cache blocks into a single unified cache, forming the sparse KV Cache produced by our \titleabbs approach.

% Top Pi是文章内的，Top K是文章间的取最好的
% 把每个文档选出来的block放到同一个列表里，根据attention score的相对压缩比例再对比

% 为每个文档计算一个percentage

% % $N^*$, as the statically important layer, is analyzed by Appendix A.2.
% After calculating the inner product based on $Q_i$ with the mean K of each block, the KVs ranked Top $P_i$ is selected.
% To further reduce the GPU memory, \titleabbs aggregates the selected middle KV Cache and filters them again proportionally according to their computed inner product scores, taking the Top K to satisfy $\text{K} = |\text{blocks}| / |\text{contexts}|$.
% So far, the final KV contains: initial, filtered, and local.
% These KVs are to be used for final inference after being spliced in their original relative order \citep{DBLP:conf/nips/XiaoZ0XLZ0024}.
% An example of the selected important KVs is shown in Figure~\ref{fig:recomputation}.

\subsection{Recomputation}

\begin{figure}[t]
    \centering
    \includegraphics[width=0.75\linewidth]{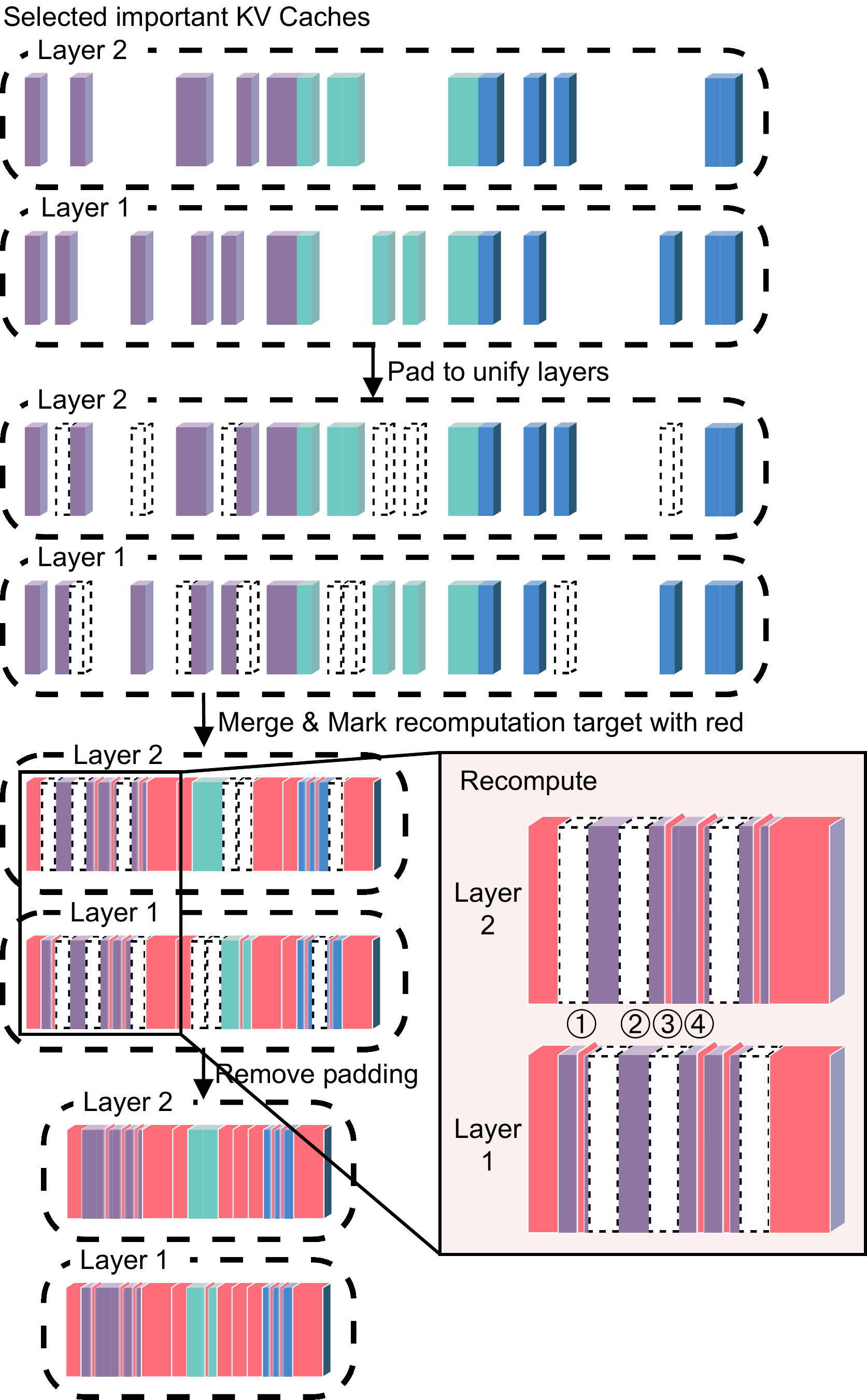}
    \caption{Case of the recomputation for layers 1 and 2. Four token recomputation examples are labels: \ding{172} lower layer recomputed but upper layer not, \ding{173} neither layer recomputed, \ding{174} upper layer recomputed but lower layer not, and \ding{175} both layers recomputed.}
    \label{fig:recomputation}
\end{figure}

Despite obtaining the sparse KV Cache in Section~\ref{sec:select}, the lack of cross-attention between KV Caches from different contexts, which are prefilled separately rather than jointly, leads to performance degradation \citep{DBLP:journals/corr/abs-2405-16444}. To mitigate this, we need to recompute some tokens in the sparse KV Caches.
In addition to tokens with high attention scores from the initial and local positions, we further incorporate tokens that exhibited significant attention weights in the original context for recomputation (see Appendix A.1 for implementation details).
However, \textbf{sparse KV Caches from multiple contexts cannot be aligned across layers}, posing a significant challenge and explaining why other works avoid sparse KV Caches for multi-context settings. As shown in Figure~\ref{fig:recomputation}, there are many misaligned blocks between layer 1 and layer 2 in the selected KV Caches.

To address the challenge of cross-layer block alignment, we first align mismatched positions using blank blocks, then apply the following recomputation rules:
\begin{enumerate}
    \item For any token requiring recomputation at layer $n$, compute outputs from all preceding $n-1$ layers.
    \item At layer $n$, recompute tokens when necessary while reusing existing Cache entries otherwise.
\end{enumerate}
Following recomputation, remove all padding blocks. Figure~\ref{fig:recomputation} illustrates this complete workflow: padding, merging, recomputing, and finally removing padding elements.

Examining the four examples in Figure~\ref{fig:recomputation} demonstrates strict adherence to the two rules.
Example \ding{172} involves recomputing only layer-1 tokens' KV pairs without output computation since layer-2 remains unchanged.
Example \ding{173} requires no computation under the rule 1.
In \ding{174} and \ding{175}, layer-2 tokens require recomputation, necessitating computation of corresponding layer-1 token outputs.
Rule 2 employs an optimization strategy to maximize reuse of existing KV Cache entries, for instance, in example \ding{174}, while layer-2 requires recomputation, layer-1 padding blocks utilize the token's prefilled KV Cache for output computation.
This approach effectively reduces computational overhead while compensating for missing cross-document attention.

% 1. 上层：有kv，并且重算，下层没kv
% 2. 上层有kv且重算，但下层不重算
% 3. 上层重算，下层有kv重算
% 4. 上层不重算，

% 图里展示4种情况，文字里两个规则

% Here we unfold the explanation for the special locations.
% In the recomputation details of Figure~\ref{fig:recomputation}, with the token in layer 2 needing to be recomputed, but in layer 1 located in the padding block (the position of $\downarrow$), calculate its Q-state but reuse its KV-state in layer 1.
% This is equivalent to passing the token from layer 2 to layer 1, corresponding to rule 1.
% And, for the token that is not recomputed at layer 2 but recomputed at layer 1 (the position of $\top$), layer 1 only computes its KV-state, matching rule 3.

% Note that only the rules for layer $n$ and layer $n+1$ are revealed here.
% Actually, this recomputation connection is not just between neighboring layers, but is passed from the top layer to the bottom.

After recomputing new KV values, conventional approaches directly overwrite the old KV Cache with the new values.
In our work, we propose two update strategies:
\begin{itemize}
    \item \textbf{Overwrite}: follow the traditional update mechanism by replacing the old Cache with new KV values;
    \item \textbf{Fusion}: combine new KV values with the old Cache.
\end{itemize}
While the overwrite mechanism is straightforward and requires no further elaboration, the fusion strategy warrants detailed discussion.

The key to KV fusion lies in determining the appropriate blending ratio between new KV values ($KV_{\text{doc-i}^\text{new}}$) and the old Cache ($KV_{\text{doc-i}^\text{old}}$), with their weights summing to 1. We employ a simple yet effective approach using the cosine similarity $\theta=\cos(KV_{\text{doc-i}^\text{new}}, KV_{\text{doc-i}^\text{old}})$. Experimental observations reveal that this similarity measure $\theta$ typically yields high values around 0.9.
To balance cache updates with preservation of historical information, we formulate the update rule as: 
\begin{align}
\small
\begin{aligned}
    KV_{\text{doc-i}^\text{new}}=\theta\times KV_{\text{doc-i}^\text{new}}+
     (1-\theta)\times KV_{\text{doc-i}^\text{old}}.
    \end{aligned}
    \label{equ:kv}
\end{align}
This equation guarantees both the update of inter-document relationships and the retention of intra-document information, achieving effective knowledge fusion.

% After computing the new KV pairs, a question arises: should we fully adopt the new values and discard the old ones?
% Here, we propose two options:
% \begin{enumerate}
%     \item \textbf{Overwrite}: the new KV pairs replace the old ones;
%     \item \textbf{Fusion}: the new and old KV pairs are combined.
% \end{enumerate}
% We employ a straightforward fusion approach, where the new KV pairs are dominant while incorporating a small portion of the old KV pairs.
% Specifically, we follow the equation: $\text{new KV}=\cos_{\text{KV}}\times\text{new KV}+(1-\cos_\text{KV})\times\text{old KV}$, where $\cos_\text{KV}=\cos(\text{new KV}, \text{old KV})$.
% Here, the similarity (generally greater than 0.9) is computed separately for the K-to-K and V-to-V pairs.
% It implies that in the computation of new inter-context cross-attention, we endeavor to retain some of the cross-attention from the original context to avoid information loss.

After concatenating all $KV_{\text{doc-i}^\text{new}}$ yields $KV_{\text{docs}^\text{new}}$, answer inference can proceed. However, both $Q_{\text{que}}$ and $\hat{Q}_{\text{doc-i}}$ are generated or extended based solely on the initial and local parts of the KV Caches, lacking information from the middle section. Thus, neither is suitable for answer generation. In contrast, $KV_{\text{docs}^\text{new}}$ encompasses information from initial, middle, and local sections. Therefore, we re-perform an incremental prefill of the user query based on $KV_{\text{docs}^\text{new}}$ and then infer the answer.

\section{Evaluation}

\subsection{Setup}

\begin{figure}[t]
    \centering
    \includegraphics[width=\linewidth]{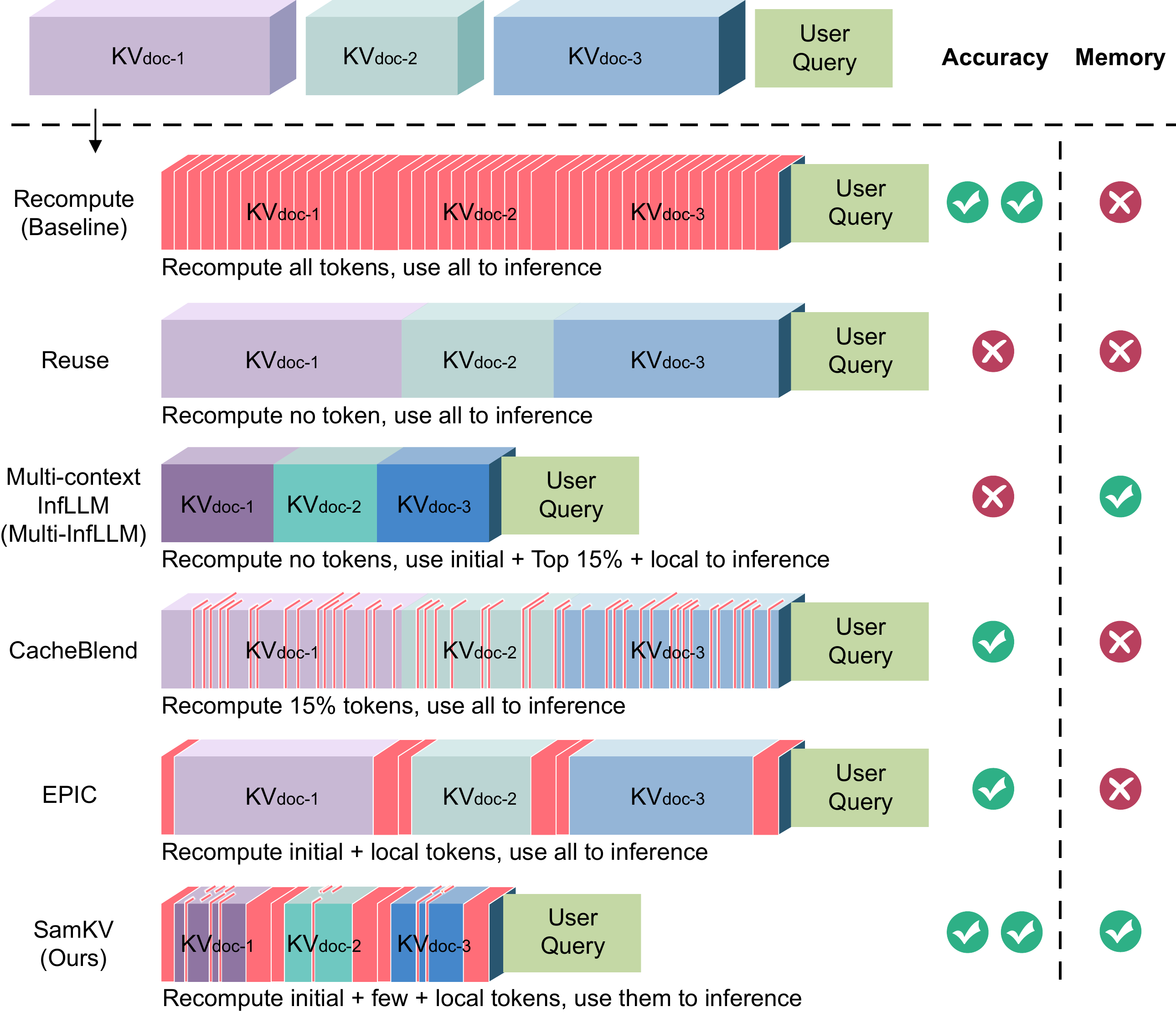}
    \caption{Comparison among multi-context methods. Darker colors indicate KV Caches obtained through sparsification. Red markings denote recomputed positions.}
    \label{fig:compare}
\end{figure}

\begin{table}[t]
\small
    \centering
    \begin{tabular}{llll}
    \toprule
        \textbf{Method} & \textbf{2WikiMQA} & \textbf{MuSiQue} & \textbf{HotpotQA} \\
    \hline
    \rowcolor{gray!20} \multicolumn{2}{l}{\textbf{\quad\emph{Mistral 7B Instruct}}}
         & &  \\
        Recompute & 25.06 & 19.94 & 37.70 \\
        Reuse & \down{6.33}{18.73} & \down{1.92}{18.02} & \down{2.32}{35.38} \\
        Multi-InfLLM & \down{23.47}{1.59} & \down{15.77}{4.17} & \up{38.61}{0.91} \\
        CacheBlend & \down{20.42}{4.64} & \down{17.44}{2.5\quad} & \down{33.98}{3.75} \\
        EPIC & \down{19.40}{5.66} & \down{17.22}{2.72} & \down{32.23}{5.47} \\
        \textbf{\titleabb-overwrite} & \up{27.04}{1.98} & \down{17.20}{2.74} & \up{\textbf{39.10}}{1.4} \\
        % \qquad - (none) & \down{20.66}{4.4} \\
        \textbf{\titleabb-fusion} & \up{\textbf{27.88}}{2.82} & \down{\textbf{18.54}}{1.4} & \down{37.66}{0.04} \\
    \hline
        \rowcolor{gray!20} \multicolumn{2}{l}{\textbf{\quad\emph{Llama 3.1 8B Instruct}}} & & \\
        Recompute & 21.48 & 14.56 & 24.12 \\
        Reuse & \down{3.66}{17.82} & \down{2.23}{12.33} & \down{2.80}{21.32} \\
        Multi-InfLLM & \down{16.20}{5.28} & \down{12.70}{1.86} & \up{29.04}{4.92} \\
        CacheBlend & \down{19.12}{2.36} & \down{12.33}{2.23} & \down{22.18}{1.94} \\
        EPIC & \down{17.36}{4.12} & \down{13.01}{1.55} & \down{23.04}{1.08} \\
        % \qquad - (none) \\
        \textbf{\titleabb-overwrite} & \up{\textbf{23.13}}{1.65} & \down{\textbf{13.04}}{1.52} & \up{\textbf{35.27}}{11.15} \\
        \textbf{\titleabb-fusion} & \up{22.41}{0.93} & \down{12.47}{2.09} & \up{30.52}{6.4}\\
    \bottomrule
    \end{tabular}
    \caption{F1 scores of cross-context KV Cache methods.}
    \label{tab:overview_result}
\end{table}

\begin{table*}[th]
\small
    \centering
    \begin{tabular}{c|c|ccc|llll|l}
    % \toprule
    % 
    \hline
     \# & \textbf{Method} & \multicolumn{3}{c|}{\textbf{Condition}} & \multicolumn{4}{c|}{\textbf{Dataset}} & \textbf{Avg.} \\
         & & \textbf{Selection} & \textbf{PersBias.} & \textbf{Recompute} & \textbf{2WikiMQA} & \textbf{MuSiQue} & \textbf{HotpotQA} & \textbf{DuReader} & \\
    \hline
    \rowcolor{gray!20} \multicolumn{9}{l}{\textbf{\quad\emph{Qwen2.5 3B Instruct}}} & \\  
    1 & Recompute & - & - & - & 33.34 & 20.96 & 40.11 & 16.49 & 27.73 \\
    2 & \textbf{\titleabb} & \cha & - & \cha & 21.86 & 13.18 & 31.73 & 16.30 & 20.77 \\
        3 & & \cha & - & \go & 34.01 & 16.94 & 36.51 & 17.70 & 26.29 \\
        4 & & \go & \cha & \cha & 22.29 & 10.75 & 26.22 & 15.00 & 18.57 \\
        5 & & \go & \go & \cha & 23.64 & 12.76 & 28.52 & 14.90 & 19.96 \\
        6 & & \go & \cha & \go & 33.10 & 13.18 & 38.57 & 17.82 & 25.67 \\
        7 & & \go & \go & \go & \up{\textbf{35.00}}{1.66} & \down{\textbf{17.03}}{3.93} & \down{\textbf{39.09}}{1.02} & \up{\textbf{18.08}}{1.59} & \down{\textbf{27.30}}{0.43} \\
    \hline
    \rowcolor{gray!20} \multicolumn{9}{l}{\textbf{\quad\emph{Llama 3.1 8B Instruct}}} & \\  
    8 & Recompute & - & - & - & 21.48 & 14.56 & 24.12 & 27.60 & 21.94 \\
    9 & \textbf{\titleabb} & \cha & - & \cha & 10.54 & 9.49 & 21.68 & 16.99 & 14.68 \\
        10 & & \cha & - & \go & 22.12 & 12.44 & 26.67 & 25.00 & 21.56 \\
        11 & & \go & \cha & \cha & 10.63 & 9.59 & 23.62 & 16.14 & 15.00 \\
        12 & & \go & \go & \cha & 10.91 & 9.36 & 23.60 & 17.74 & 15.40 \\
        13 & & \go & \cha & \go & 20.90 & \down{\textbf{13.21}}{1.35} & 29.26 & 25.48 & 22.21 \\
        14 & & \go & \go & \go & \up{\textbf{22.41}}{0.93} & 12.47 & \up{\textbf{30.52}}{6.4} & \down{\textbf{25.58}}{2.02} & \up{\textbf{22.75}}{0.81} \\
    \hline
    % \bottomrule
    \end{tabular}
    \caption{Using full recomputation as the baseline, ablations are conducted on three aspects: selecting the KV Cache in the middle section, adding personalized bias (PersBias.), and whether to recompute (with fusion as default). The evaluation metric is the F1 score, with Avg. denoting the average F1 score across datasets.}
    \label{tab:our_result}
\end{table*}

Our proposed \titleabbs is fundamentally designed as a sparse processing strategy for multi-context scenarios.
Given this intrinsic characteristic, our experimental setup including the dataset selection, base LLM configuration, and baseline comparison aligns with established practices in multi-context research. Specifically, we adopt the evaluation framework used by existing multi-context methods such as CacheBlend \citep{DBLP:journals/corr/abs-2405-16444} and EPIC \citep{DBLP:journals/corr/abs-2410-15332} to ensure fair and meaningful comparisons.

\textbf{Implementation.}
We implement \titleabbs by modifying the original Transformers library's model code \citep{DBLP:conf/emnlp/WolfDSCDMCRLFDS20}.
Compared to vLLM-based implementations, our approach requires fewer code modifications and demonstrates better compatibility with framework upgrades.
For inference hyperparameters, we employ greedy decoding with temperature set to 0.
Regarding KV Cache management at the block level, we configure \titleabbs with a block size of 64, allocating 1 block for the initial position KV Cache and 2 blocks for the local position KV Cache during sparsification.

\textbf{Datasets.}
For fair comparison with existing methods, we evaluate performance on three multi-context QA datasets from LongBench \cite{DBLP:conf/acl/BaiLZL0HDLZHDTL24}: 2WikiMQA, MuSiQue, and HotpotQA. In addition to these datasets, we conduct further ablation experiments on DuReader.
Each dataset contains 200 test samples, with each sample comprising a long context, user query, and corresponding answer.
% We employ dataset-specific delimiters (e.g., ``Passage:'' for MuSiQue) to split a long context into multiple segments.
All answers are uniformly evaluated using F1 scores.
% In these QA tasks, all answers are uniformly evaluated using F1 scores.

\textbf{Models.}
In our experiments, we select several representative models including Mistral 7B Instruct \citep{DBLP:journals/corr/abs-2310-06825} and Llama3.1 8B Instruct \citep{DBLP:journals/corr/abs-2407-21783} based on current hardware compatibility.
To provide comprehensive evaluation, we incorporate the Qwen2.5 3B Instruct \citep{qwen2.5} to examine performance across different model scales.

\textbf{Methods.}
We compare our \titleabbs with five alternative approaches: Recompute, Reuse, Multi-context InfLLM (Multi-InfLLM), CacheBlend \citep{DBLP:journals/corr/abs-2405-16444}, and EPIC \citep{DBLP:journals/corr/abs-2410-15332}. Among these, Recompute serves as our baseline by performing complete recalculation of the KV Caches.
% All other methods are evaluated relative to Recompute's performance.
Multi-InfLLM concatenates KV Caches from all contexts into a single context Cache and applies InfLLM's single-context sparsification method \citep{DBLP:conf/nips/XiaoZ0XLZ0024}.

\textbf{Ablation studies.}
The ablation study examines three key aspects: selection of KV Caches, personalized bias, and recomputation. For the selection component, the sparse KV Caches with selection incorporate the initial-position KV Cache, selected important KV Caches, and local-position KV Caches. In contrast, the no-selection variant only utilizes the initial and local KV Caches without selection.

% init+local的先省略掉，篇幅过长了，没有地方写结果和分析了
% To verify the effectiveness of KV Cache sparsification for middle segments, we include an additional configuration where the KV Caches only contains initial and local segments from all contexts, completely excluding middle segments. This configuration is labeled as Initial+Local in our experiments.

\textbf{Environment.}
We conduct experiments on a server equipped with a 64GB NPU and a 256-core Kunpeng 920 CPU@2.60GHz (hyperthreading disabled), supported by 2TB DRAM. The system runs EulerOS 2.0 with kernel version 5.10.0. To ensure fair comparison across different hardware platforms, we maintain F1 score comparisons when evaluating against CacheBlend and EPIC.
% For all other experiments, we continue reporting both TTFT and F1 metrics.

\subsection{Overall improvement}
\label{sec:overall_improvement}

From Table~\ref{tab:overview_result}, it is evident that \titleabbs outperforms all other multi-context methods and even surpasses Recompute, the baseline, on both 2WikiMQA and HotpotQA.
Even better, when based on Llama, SamKV-overwrite achieves an F1 score of 35.27 on HotpotQA, significantly surpassing Recompute's 24.12, which we attribute to the inclusion of excessive irrelevant information in full recomputation that interferes with result generation.
% This suggests that sparsifying the multi-context KV Caches does not degrade performance while reducing GPU memory usage.
Furthermore, the performance of overwrite and fusion strategies varies depending on the model and dataset, with each showing different advantages. Overall, both recomputation methods contribute to improved effectiveness.

\subsection{Ablation studies}

We conduct comprehensive ablation experiments to evaluate three key design choices: (i) whether to apply selection to the middle segment’s KV Cache, (ii) the inclusion of personalized bias, and (iii) the use of recomputation. Following the comparative analysis of overwrite versus fusion recomputation strategies in Section~\ref{sec:overall_improvement}, we fix the recomputation method to fusion for all ablation trials to isolate variable effects.
The detailed analysis is presented as follows.

\textbf{Selection of middle KV Caches.}
The key consideration in adopting selection lies in whether to extract crucial KV from the middle part of the KV Cache. Here, \go~denotes selection, while \cha~indicates no selection, meaning only the initial and local parts of the KV Caches are used.
Comparing the average results (Avg.) in Table~\ref{tab:our_result}, specifically rows 2 and 4 as well as rows 9 and 11, it can be observed that when recomputation is disabled, the absence of the middle KV Cache yields slightly better performance than its inclusion.
This suggests that relying solely on the initial and local parts of the KV Cache can achieve reasonably good results.
However, this performance improvement precisely becomes the factor that limits its further enhancement.
Only when extending to incorporate the middle part can additional personalized bias be introduced. For instance, when recomputation is enabled, comparing the cases without selection (rows 3 and 10) and those with personalized bias (rows 7 and 14) reveals that selection with personalized bias delivers superior performance. 
Evidently, adopting selection is a prerequisite for further enhancement.
The computational overhead associated with selection is discussed in detail later.

\textbf{Personalized bias.}
Comparing rows 4 and 5, rows 6 and 7, rows 11 and 12, and rows 13 and 14 in Table~\ref{tab:our_result} reveals that while both cases involve selection, the inclusion of personalized bias in the query vectors yields better performance.
The results show that introducing a small amount of Q Cache into the query vector may help identify consensus for better performance.
% The results demonstrate that introducing a small amount of Q Cache from other contexts during the search process enhances the retrieval of consensus information across documents. This consensus information further contributes to improving answer accuracy.
% 添加个性化偏置的这个是否有利于找到文档间共识部分，如果想证明的话，需要更进一步的实验：两种，一种是包含bias，一种不包含bias，两种都去稀疏化，看稀疏化的内容与其他文档的相关性，统计为具体数值，以体现其提升（这里没法加入进一步的实验了）

\textbf{Recomputation.}
Comparing rows 2 and 3, rows 4 and 6, rows 5 and 7, rows 9 and 10, rows 11 and 13, and rows 12 and 14 in Table~\ref{tab:our_result}, it can be observed that recomputing only a portion of the sparsified KV Cache consistently improves the F1 score by 6\% to 7\%, regardless of whether selection is applied or personalized bias is incorporated.
This indicates that recomputing only the sparse subset of tokens can significantly alleviate the missing cross-attention between documents without requiring the full KV Cache to be carried throughout the entire inference process.

In terms of time consumption, compared to methods like CacheBlend and EPIC that utilize full KV Caches, our approach introduces an additional sparsification process which incurs extra time overhead. However, these methods require carrying the complete KV Caches during recomputation, for example, when recomputing the KV of the local position in $n$-th layer, the output of the local position in the $(n-1)$-th layer relies on the KV of all tokens for computation, resulting in high time complexity. In contrast, our method only recomputes a small subset of tokens from the sparsified KV Caches, and the sparsification process is accelerated by vector databases and GPUs, leading to lower overall time consumption.
%than CacheBlend and EPIC.
% In terms of time consumption, compared to multi-context methods such as CacheBlend and EPIC which require recomputing with the full KV Cache, our approach introduces an additional sparsification step but with only a small portion of the KV Cache obtained through sparsification recomputed, the overall time cost is reduced. Moreover, the sparsification process employs traditional vector inner products, which are compatible with vector databases and GPU acceleration, resulting in slightly lower time consumption than CacheBlend and EPIC.

\section{Conclusion}

In this paper, we present \titleabb, the first method designed for KV Cache sparsification across diverse contexts. Experimental results demonstrate that \titleabbs achieves comparable accuracy to full recomputation while reducing the required KV Cache during inference to only 15\% of its original length. Further investigations reveal that incorporating inter-document consensus during sparsification along with localized recomputation after sparsification not only reduces computational overhead and the amount of KV Cache needed for inference but also significantly mitigates cross-attention deficiencies between documents. These findings provide valuable insights for optimizing ultra-long multiple contexts processing and establish an important reference for future research in this domain.

\begin{figure*}[t]
    \centering
    \includegraphics[width=0.9\linewidth]{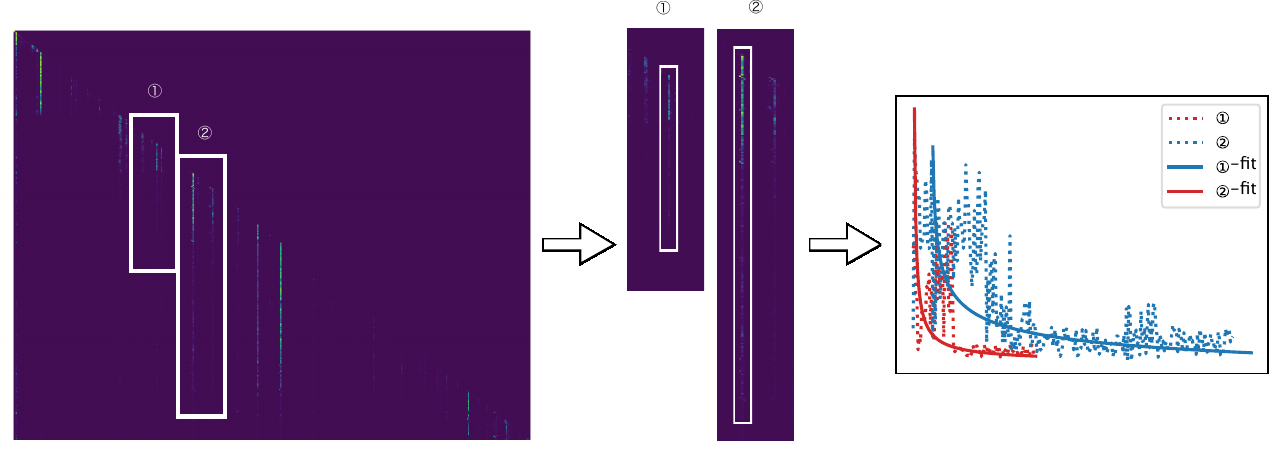}
    \caption{Attention sink (left) in the middle of the context \citep{hu2025raas} evaluated on the MATH 500 \citep{DBLP:conf/nips/HendrycksBKABTS21} using the Qwen2.5 Math 7B Instruct \citep{DBLP:journals/corr/abs-2409-12122}. Extract representative tokens within the block (middle), and plot them as a dashed figure (right). Fit the dashed line with a power law to get the solid line.}
    \label{fig:attention_matrix}
\end{figure*}
\begin{figure*}[t]
    \centering
    \includegraphics[width=0.9\linewidth]{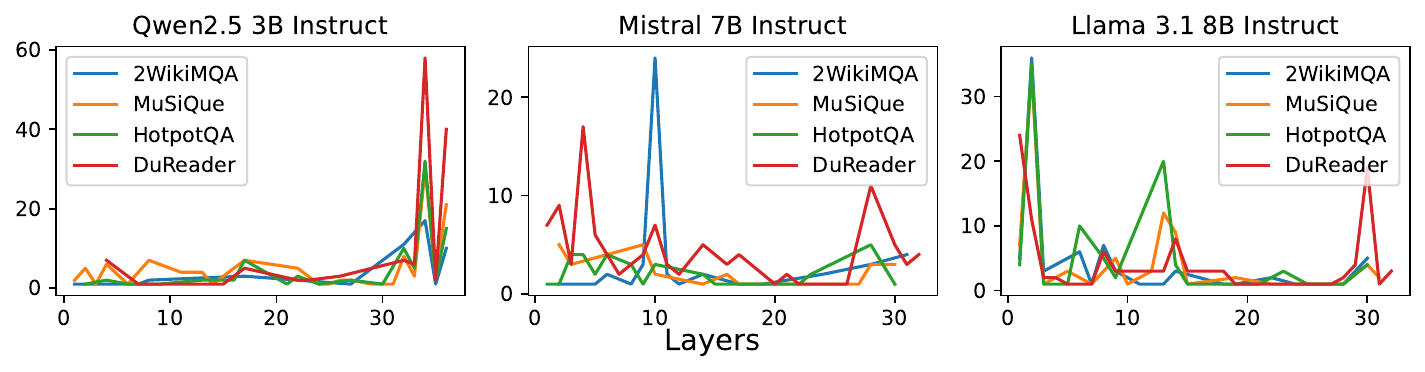}
    \caption{After analyzing block importance across datasets, \titleabbs evaluates the stability of attention distributions in each layer. The vertical axis measures the stability of each layer's attention allocation to the most important block. Higher stability values indicate more consistent attention distributions in the corresponding layer.}
    \label{fig:layers}
\end{figure*}

\appendix

\section{Static attention analysis of KV Cache}

\subsection{Attention for blocks within a layer}
\label{app:block_importance}

When managing KV Cache at the block level for long contexts, a block's attention scores inherently depend on the relative attention weights of its internal tokens within the full sequence. Conventional analysis interprets these scores unidirectionally: higher scores indicate greater importance, while lower scores suggest lesser importance. In contrast to this single-score representation, we propose a dual-score characterization for each block, explicitly decoupling and quantifying both its importance and unimportance attributes.

\textbf{Importance attribute.}
Since a block contains numerous tokens, computing relative attention between all tokens would introduce excessive complexity. Instead, we select the most representative token from each block to determine its importance attribute, then compare attributes across blocks. Specifically, for each block, we identify the token that receives consistently high attention from subsequent tokens, visualized as distinct bright lines in Figure~\ref{fig:attention_matrix} (left). Extracting these lines yields Figure~\ref{fig:attention_matrix} (middle), and plotting their values produces the dashed curves in Figure~\ref{fig:attention_matrix} (right). These curves closely follow a power-law distribution ($y\propto x^{-\alpha}, \alpha>0$), approximated by the solid lines in Figure~\ref{fig:attention_matrix} (right).
The fitted curves reveal two key observations: a smaller exponent $\alpha$ corresponds to higher overall attention (e.g., \ding{172}-fit), while a larger $\alpha$ indicates lower attention (e.g., \ding{173}-fit). Thus, block importance is derived by sorting their respective $\alpha$ values. Moreover, since attention reflects token significance, we apply the PauTa Criterion to detect outliers in $\alpha$s, identifying tokens requiring recomputation in the middle KV Cache section.

\textbf{Unimportance attribute.}
Unlike important tokens that exhibit distinct bright lines in attention patterns, unimportant blocks or tokens consistently show uniformly low attention values, appearing as large purple regions in Figure~\ref{fig:attention_matrix} (left). To quantify this, we select the token with the highest sustained attention within each block and compute its mean attention score. A lower mean indicates weaker attention allocation. When even the most prominent token in a block receives low attention, the block itself is deemed less important, e.g., its unimportance is more pronounced. Based on this measure, we derive an unimportance ranking for all blocks within a layer.

By extracting the block with the highest importance attribute and the block with the highest unimportance attribute from the KV Cache of document i, we obtain $KV^{(n)}_{\text{doc-i}_\text{max}}$ and $KV^{(n)}_{\text{doc-i}_\text{min}}$, which are used to compute $P^{(n)}_\text{doc-i}$.

\subsection{Block importance cross layers}
\label{app:attn_sink_layer}

Through manual inspection of attention score maps from the layers, we observe that tokens with high attention scores (e.g., those exhibiting distinct bright lines in Figure~\ref{fig:attention_matrix}) consistently maintain high attention across multiple layers. For instance, in Qwen2.5 3B Instruct, the last five layers all highlight tokens at nearly identical positions. This indicates strong consistency in high-attention token selection across some layers.
% As the model approaches its output, the locations of high-attention tokens stabilize significantly.

To better observe which layers maintain stable attention distributions, we transform and visualize this process. For efficient analysis and management, we examine attention patterns at the block level. Appendix~\ref{app:block_importance} provides the attention ranking for all blocks in each layer. For any given block, we average its rankings across all layers to determine its overall attention ranking in the model. Our key insight is that \textbf{when a block achieves the highest attention in the full model and certain layers consistently rank it highest, these layers demonstrate consensus regarding the block's attention stability}. We identify such layers as having stable attention patterns.

We implement this insight through a two-step pipeline. First, we identify the block $\beta$ with maximum attention using the method described in Appendix~\ref{app:block_importance}. Second, if $\beta$'s ranking in layer $n$ is statistically significant as determined by the PauTa Criterion (indicating an outlier), we classify this layer as attention-stable and increment its score by 1. Figure~\ref{fig:layers} visualizes the resulting scores across all layers as a line chart.

The figure reveals performance variations across models and datasets, demonstrating that this attention stabilization phenomenon is model dependent.
Observations of attention stability distributions across different models reveal that higher stability is consistently exhibited in the final few layers. Therefore, we focus on selecting from these latter layers.
Specifically, the phenomenon occurs in:
\begin{itemize}
    \item Qwen2.5-3B-Instruct layers 32 to 36;
    \item Mistral-7B-Instruct layers 28 to 32;
    \item Llama-3.1-8B-Instruct layers 29 to 32.
\end{itemize}
We identify these layers as $N^*$ due to their stable attention distributions and high consistency. From these layers, we extract three key components: the anchor block, the maximum-attention block, and the minimum-attention block. These components are used to compute $P^{(n)}_{\text{doc-i}}$, which yields both stable and reasonable sparsity ratios.

\bibliography{aaai2026}

\end{document}